\documentclass{article}
\usepackage{arxiv}

\usepackage[utf8]{inputenc} % allow utf-8 input
\usepackage[T1]{fontenc}    % use 8-bit T1 fonts
\usepackage{hyperref}       % hyperlinks
\usepackage{url}            % simple URL typesetting
\usepackage{booktabs}       % professional-quality tables
\usepackage{amsfonts}       % blackboard math symbols
\usepackage{nicefrac}       % compact symbols for 1/2, etc.
\usepackage{microtype}      % microtypography
\usepackage{lipsum}		% Can be removed after putting your text content
\usepackage{graphicx}
\usepackage{natbib}
\usepackage{doi}
\usepackage{times} % assumes new font selection scheme installed
\usepackage{amsmath}
\usepackage{amssymb}  % assumes amsmath package installed
\usepackage{booktabs}
\usepackage{multirow}
\usepackage{comment}
\usepackage{float}
\usepackage{subfig}
\usepackage{caption}
\usepackage[export]{adjustbox}

\usepackage{enumitem}
\DeclareMathOperator*{\argmax}{arg\,max}

\title{Adaptive Social Force Window Planner with Reinforcement Learning}

\author{
 Mauro Martini \\
  Department of Electronics and Telecommunications \\
  Politecnico di Torino\\
  Torino, TO, 10129, Italy \\
  \texttt{mauro.martini@polito.it} \\
   \And
 No\'e P\'erez-Higueras \\
  School of Engineering \\
  Pablo de Olavide University\\
  Crta. Utrera km 1, Seville, Spain \\
  \texttt{noeperez@upo.es} \\
   \And
 Andrea Ostuni \\
  Department of Electronics and Telecommunications \\
  Politecnico di Torino\\
  Torino, TO, 10129, Italy \\
  \texttt{andrea.ostuni@polito.it} \\
  \And
 Marcello Chiaberge \\
  Department of Electronics and Telecommunications \\
  Politecnico di Torino\\
  Torino, TO, 10129, Italy \\
  \texttt{marcello.chiaberge@polito.it} \\
  \And
 Fernando Caballero \\
  School of Engineering \\
  Pablo de Olavide University\\
  Crta. Utrera km 1, Seville, Spain \\
  \texttt{fcaballero@upo.es} \\
  \And
 Luis Merino \\
  School of Engineering \\
  Pablo de Olavide University\\
  Crta. Utrera km 1, Seville, Spain \\
  \texttt{lmercab@upo.es} \\
}
\renewcommand{\shorttitle}{\textit{arXiv} Template}

\hypersetup{
pdftitle={Adaptive Social Force Window Planner with Reinforcement Learning},
pdfsubject={adaptive-social-navigation-preprint},
pdfauthor={Mauro Martini},
pdfkeywords={Human-Aware Motion Planning, Reinforcement Learning, Service Robotics},
}

\begin{document}
\maketitle

\begin{abstract}

Human-aware navigation is a complex task for mobile robots, requiring an autonomous navigation system capable of achieving efficient path planning together with socially compliant behaviors. Social planners usually add costs or constraints to the objective function, leading to intricate tuning processes or tailoring the solution to the specific social scenario. Machine Learning can enhance planners' versatility and help them learn complex social behaviors from data. This work proposes an adaptive social planner, using a Deep Reinforcement Learning agent to dynamically adjust the weighting parameters of the cost function used to evaluate trajectories. The resulting planner combines the robustness of the classic Dynamic Window Approach, integrated with a social cost based on the Social Force Model, and the flexibility of learning methods to boost the overall performance on social navigation tasks.
Our extensive experimentation on different environments demonstrates the general advantage of the proposed method over static cost planners.
\end{abstract}


\section{Introduction}
\label{sec:intro}
In recent years, service robots have emerged as a promising automation solution in various social contexts, ranging from domestic assistance \cite{eirale2022human, eirale2022marvin} to health-care \cite{holland2021service}. These advancements have opened up new avenues for robotics research, particularly in human-aware navigation. The robotics community has proposed different benchmarks to evaluate the existing social navigation algorithms \cite{khambhaita2017assessing, PerezRal2023}.

\begin{figure}[ht]
\centering
\includegraphics[width=0.55\linewidth]{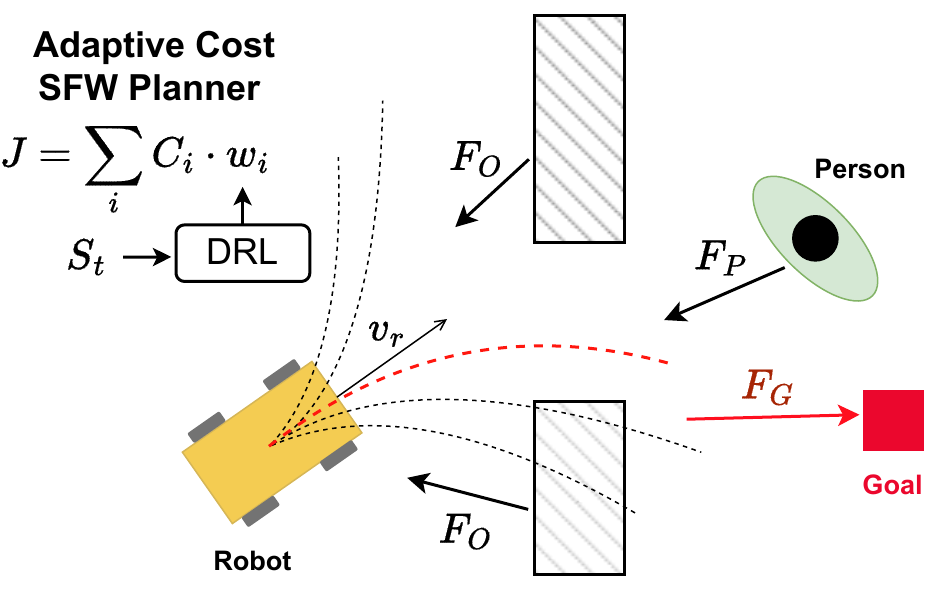}
\caption{The Social Force Window (SFW) Planner combines standard Dynamic Window Approach and Social Force Model. The trajectory scoring process is optimized by a DRL agent that dynamically adjust the cost weights based on local environmental conditions.}
\label{fig:first_page}
\end{figure}

However, social navigation is a complex problem that poses contrasting objectives and is often difficult to formulate with an analytical expression, as is usually done in classic navigation cost functions. This complexity arises from the intricate dynamics of human behavior and the multitude of social rules not considered in standard path planning. Different social navigation scenarios have been partially categorized in the literature to build consistent research and benchmarks \cite{gao2022evaluation, mavrogiannis2023core}, highlighting unique challenges in each situation. Standard social planners struggle to perform properly in all of them, considering that environmental geometry and features are crucial in constraining navigation in cluttered, narrow passages or wide open spaces. Therefore, the diversity and unpredictability of social scenarios necessitate a more flexible and adaptive approach. 

In this context, Machine Learning (ML) techniques represent a potential solution to this problem. ML models can leverage data to learn behaviors that enhance mobile robots' adaptability to new situations\cite{xiao2020motion} without being explicitly programmed for a specific task. Among existing ML paradigms, Deep Reinforcement Learning (DRL) is particularly suited for learning behavioral policies and, hence, for navigation \cite{zhu2021deep}. Recent studies tried to address the challenges posed by human-aware navigation with DRL \cite{chen2017socially}, as better discussed in Section \ref{sec:related_works}.
On the other hand, end-to-end learning approaches may often present less robust performance and poor generalization to unseen testing conditions. The authors in \cite{xu2021machine} have proposed a precious comparison between end-to-end and parameter-learning approaches, highlighting the improved performance of hybrid solutions combining standard controllers and learning.

This work proposes an adaptive parameter-learning approach for social navigation. A social controller is designed by adding a social cost to the Dynamic Window Approach (DWA). This cost is computed considering the robot-pedestrian interaction according to the Social Force Model (SFM) \cite{helbing1995social, moussaid2010walking}. The proposed solution leverages the advantage of a DRL agent to dynamically adapt the cost weights of the  human-aware local planner to different social scenarios. . 
From a general perspective, this research aims to enhance the performance and versatility of service mobile robots in general social contexts. The contribution of this paper can be summarized in (i) a social-aware local planner based on SFM, used as a baseline solution, that we refer to as Social Force Window (SFW) planner; (ii) an adaptive cost optimization of the SFW planner with a DRL agent, managing the cost terms dynamically according to the context.

\section{Related Works}
\label{sec:related_works}
This section presents and discusses the landscape of existing adaptive navigation solutions for general and social scenarios, pointing out Deep learning-driven methods.
Diverse end-to-end learning approaches \cite{chen2017socially, mirsky2021prevention} have been directly applied to navigation control. Deep Learning can also select the most suitable social navigation strategy according to the context, as done in \cite{banisetty2021deep}, which provides the robot with adaptive behavior.
Inverse Reinforcement Learning (IRL) inspired a family of recent works. It is used by \cite{kim2016socially} to learn diverse cost functions according to the social navigation scenarios. Similarly, the RTIRL \cite{ijsr18irlrrt} and PRTIRL \cite{Ding2022prtirl} use IRL to adjust the parameters of an RRT{$^*$} local path planner treated as a black box.

\begin{figure}[ht]
\centering
\includegraphics[width=\linewidth]{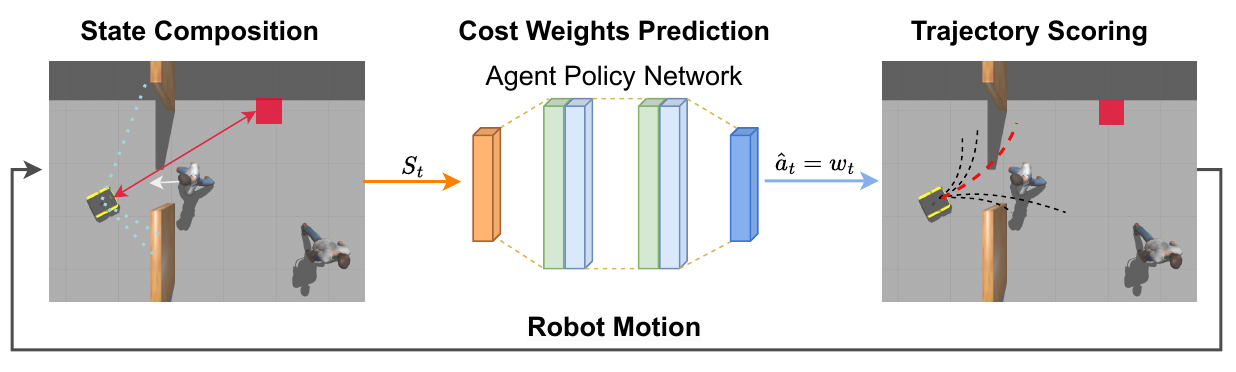}
\caption{Workflow of the main step performed by the proposed adaptive Social Force Window (SFW-SAC) Planner with DRL. The policy network learns to set the weights of the social cost used by the DWA for each situation.}
\label{fig:main_scheme}
\end{figure}

Hybrid solutions between classic controllers and learning methods have been proposed to boost the robustness of autonomous agents. A DRL approach is employed in \cite{Patel2020dwarl} to evaluate the projected trajectories of the classical Dynamic Windows Approach local planner (DWA), learning a reward function for navigating in dynamic environments. The parameter-learning presented in the family of APPL approaches (Adaptive Planner Parameter Learning) \cite{xiao2022appl} results in a promising direction to convey robustness and versatility in a unique solution \cite{xu2021machine}. APPL aims to learn a parameter management policy that can dynamically adjust the hyper-parameters of classical navigation algorithms according to the environment geometry. The authors proposed different ML techniques, including RL \cite{xu2021applr}. However, the adaptive parameter approach is applied only in static environments. 
Finally, an adaptive DWA implementation has been proposed in \cite{kobayashi2023dwaq}, dynamically changing the basic cost terms of the algorithm with a Q-table approach. Our work is an improvement and extension of these studies: we frame the adaptive control method in a social navigation problem, adopting a DRL agent working with continuous action space.

\section{Methodology}
\label{sec:methodology}
\subsection{Social Force Window Planner}
The Dynamic Window Approach (DWA) is an extremely popular local path planning method in mobile robotics \cite{fox1997dwa}. DWA generates velocity commands that comply with the robot's kinematics constraints. The search space is, therefore, restricted to velocities that can be reached quickly and avoid collisions with obstacles. %The algorithm then scores the simulated circular trajectories obtained by choosing each candidate linear and angular velocity pair $(v, w)$ for a short time interval. 
The classic objective function used to evaluate the trajectories comprises three terms associated with the goal, the velocity, and the obstacles.
A human-aware local planner has been proposed, adding a social cost to the classic DWA trajectory scoring function. For the social cost, a "social work" quantity is adopted by using the Social Force Model (SFM) of interaction between a crowd of agents proposed by \cite{helbing1995social, moussaid2009experimental, moussaid2010walking}. 
A social work $C_s$ is computed at each time step for the robot according to the following expression:
% formula social work
\begin{equation}
    C_{s} = W_r + \sum_i W_{p,i}
\end{equation}

\noindent where $W_r$ is the sum of the modulus of the robot social force ($F_P$) and the robot obstacle force ($F_O$) along the trajectory according to the SFM, while $W_p$ is the sum of the modulus of the social forces generated by the robot for each pedestrian $i$ along the trajectory. A schematic representation of forces acting on the robot is shown in Figure \ref{fig:first_page}. The goal produces an attractive force while the obstacles and pedestrians generates different repulsive forces.

The overall cost function for trajectory scoring can be formulated as:
\begin{equation}
    J = C_s \cdot w_s + C_o \cdot w_o + C_v \cdot w_v + C_d \cdot w_d + C_h \cdot w_h 
\end{equation}
\noindent where we have a single cost term for social navigation $C_s$, obstacles in the costmap $C_o$, robot velocity $C_v$, distance $C_d$ and heading $C_h$ from a local waypoint on a given global path. The costs are combined using weights $w$ that regulate the impact of each term in the velocity command selection.
We refer to this advanced social version of the DWA as a Social Force Window (SFW) planner which is publicly available in Github\footnote{\url{https://github.com/robotics-upo/social_force_window_planner}}. This local planner aims to generate safe, efficient, and human-aware paths. However, finding an optimal trade-off between all those desired aspects in every environmental context is not easy. Hence, we tackle the challenge by using a Reinforcement Learning approach to dynamically handle the weights of the costs.

\subsection{Deep Reinforcement Learning framework}
A typical Reinforcement Learning (RL) framework can be formulated as a Markov Decision Process (MDP) described by the tuple $(\mathcal{S},\mathcal{A}, \mathcal{P}, R, \gamma)$. An agent starts its interaction with the environment in an initial state $s_0$, drawn from a pre-fixed distribution $p(s_0)$ and then cyclically selects an action $\mathbf{a_t} \in \mathcal{A}$ from a generic state $\mathbf{s_t} \in \mathcal{S}$ to move into a new state $\mathbf{s_{t+1}}$ with the transition probability $\mathcal{P(\mathbf{s_{t+1}}|\mathbf{s_t},\mathbf{a_t})}$, receiving a reward $r_t = R(\mathbf{s_t},\mathbf{a_t})$.

In RL, a parametric policy $\pi_\theta$ describes the agent behavior. In the context of autonomous navigation, we usually model the MDP with an episodic structure with maximum time steps $T$. Hence, the agent's policy is trained to maximize the cumulative expected reward $\mathbb{E}_{\tau\sim\pi} \sum_{t=0}^{T} \gamma^t r_t$ over each episode, where $\gamma \in [0,1)$ is the discount factor. More in detail, we aim at obtaining the optimal policy $\pi^*_\theta$ with parameters $\mathbf{\theta}$ through the maximization of the discounted term:
\begin{equation}
    \pi^*_\theta = \argmax_{\pi} \mathbb{E}_{\tau\sim\pi} \displaystyle \sum_{t=0}^{T} \gamma^t [r_t + \alpha \mathcal{H}(\pi(\cdot|s_t))]
\end{equation}

\noindent where $\mathcal{H}(\pi(\cdot|s_t))$ is the entropy term, which increases robustness to noise through exploration, and $\alpha$ is the temperature parameter which regulates the trade-off between reward optimization and policy stochasticity. 

In this work, a parameter-learning approach has been adopted to develop an adaptive social navigation system. The DRL agent learns a policy to dynamically set the weights of the cost function that governs the robot's control algorithm. In particular, a Soft Actor-Critic (SAC)\cite{haarnoja2018soft} off-policy algorithm has been used to train the agent in simulation.

\subsection{SFM Adaptive Cost Approach}
The key idea of the proposed method lies in learning an optimal policy to dynamically set the weights of each objective function term used by the SFM local planner to score the simulated circular trajectories and select the next velocity command $(v,w)$. A DRL agent is trained to learn such policy given the local features of the surrounding environment and induce the local planner to choose optimal velocity commands. DRL is considered a competitive approach to tackle this complex task since it is not straightforward for a human to find an optimal trade-off between all the cost terms of a social controller in each situation. On the other hand, the overall methodology represents a robust hybrid solution that efficiently integrates the flexibility of the DRL agent policy with the reliability of a classical navigation algorithm. Moreover, the agent allows the planner to extend its adaptability to different social scenarios by learning the map between task-related and perception data to suitable cost weights. Figure \ref{fig:main_scheme} shows the main working steps of the proposed methodology.

\subsection{Reward function}
Reward shaping is a fundamental and controversial practice in model-free RL. A specific reward function, similar to the cost function of the SFW planner, has been designed to let the agent learn an optimal cost weights regulation policy among all the desired components of the overall navigation behavior.

\begin{description}[leftmargin=0pt]
% distance reward
\item[Goal distance] First, a distance advancement reward term is defined to encourage the approach of the next local goal on the global path, always placed at $2 m$ from the robot's actual pose:
\begin{equation}
    r_{d} = d_{t-1} - d_t 
\end{equation}

\noindent where $d_{t-1}$ and $d_t$ are Euclidean distances between the robot and the local goal. Local goal and final goal coincide in the final section of the trajectory. 

\item[Path alignment] Then, we define a reward contribution $r_h$ to keep the robot oriented towards the next local goal:
\begin{equation}
r_h = \left( 1 - 2 \sqrt{\left| \frac{\phi}{\pi} \right|} \right)
\end{equation}

\noindent where $\phi$ is the heading angle of the robot, namely the angle between its linear velocity and local goal on the global path.

\item[Robot velocity] A velocity reward is defined to promote faster motion when allowed by the environment:
\begin{equation}
r_v = \frac{v - v_{max}}{v_{max}}
\end{equation}

\item[Obstacle avoidance] An obstacle reward is included to encourage safe trajectory scoring and avoid collisions:
\begin{equation}
r_o = \frac{d_{o, min} - lidar_{max}}{lidar_{max}}
\end{equation}

\noindent where $d_{o, min}$ is the lowest distance measured by the LiDAR ranges at the current time step and $lidar_{max}$ is the saturation distance of LiDAR points, which is set to $3 m$ to perceive only local environmental features.

\item[Social penalty] The main reward contribution has been assigned to provide the agent with a socially compliant navigation policy. In particular, two different social terms have been considered: proxemics-based reward and social work.
The proxemics term penalizes the robot when intruding into the personal space of a person:

\begin{equation}
r_p = \frac{1}{d_{p, min}}
\end{equation}

\noindent where $d_{p, min}$ is the minimum person distance from the robot.
The social work generated by robot and people interaction according to the Social Force Model has been used as reward $r_s$, as done in the cost of the SFW planner. 

We also include a reward contribution for end-of-episode states, assigning $r_c = -400$ if a collision occurs.
The final reward signal is finally obtained linearly, combining the described terms. More in detail, $c_{d}=10.0$, $c_{h}=0.4$, $c_{o}=2.0$, $c_{p}=2.0$ and $c_{s}=2.5$ are the numerical coefficients chosen to balance the diverse reward contributions in the final signal.

\end{description}

\begin{figure}[ht]
\centering
\includegraphics[width=0.55\linewidth]{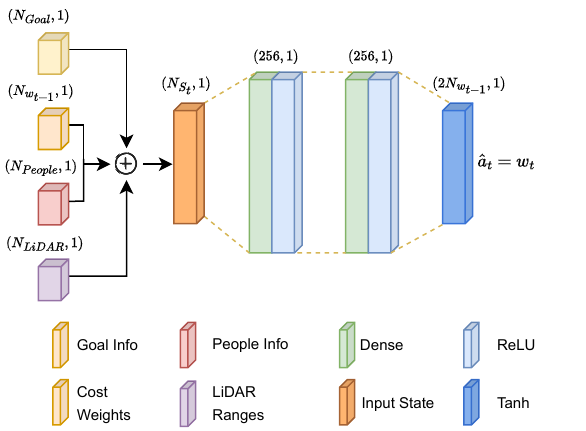}
\caption{Schematic of the policy network architecture. State composition is illustrated with separate inputs: goal distance and angle, previous cost weights, people position and velocity, and LiDAR ranges. The new cost weights are predicted as output action of the policy network.}
\label{fig:network}

\end{figure}

\subsection{Policy Neural Network and Training Design}
We define the parametrized agent policy with a deep neural network. We train the agent with the Soft Actor-Critic (SAC) algorithm presented in \cite{haarnoja2018soft}, which allows for a continuous action space and a fast convergence. In particular, we instantiate a stochastic Gaussian policy for the actor and two Q-networks for the critics.
The neural network architecture of the agent, represented in Figure \ref{fig:network}, is composed of two dense layers of $256$ neurons each. A random initial exploration phase has been performed. Random actions are then sampled with a probability that is exponentially reduced with the increase of episodes to maintain a proper rate of exploration during the whole training. Moreover, SAC uses a stochastic Gaussian policy that outputs the mean and the standard deviation of each action distribution, which are used to sample the action value at the training phase. Differently, the mean value of actions' distribution is directly used at test time. The critic networks' structure presents no differences, except they include the predicted action vector in the inputs and predict the $Q$ values.

\begin{figure}[ht]
    \centering
    {
        \includegraphics[width=0.56\linewidth]{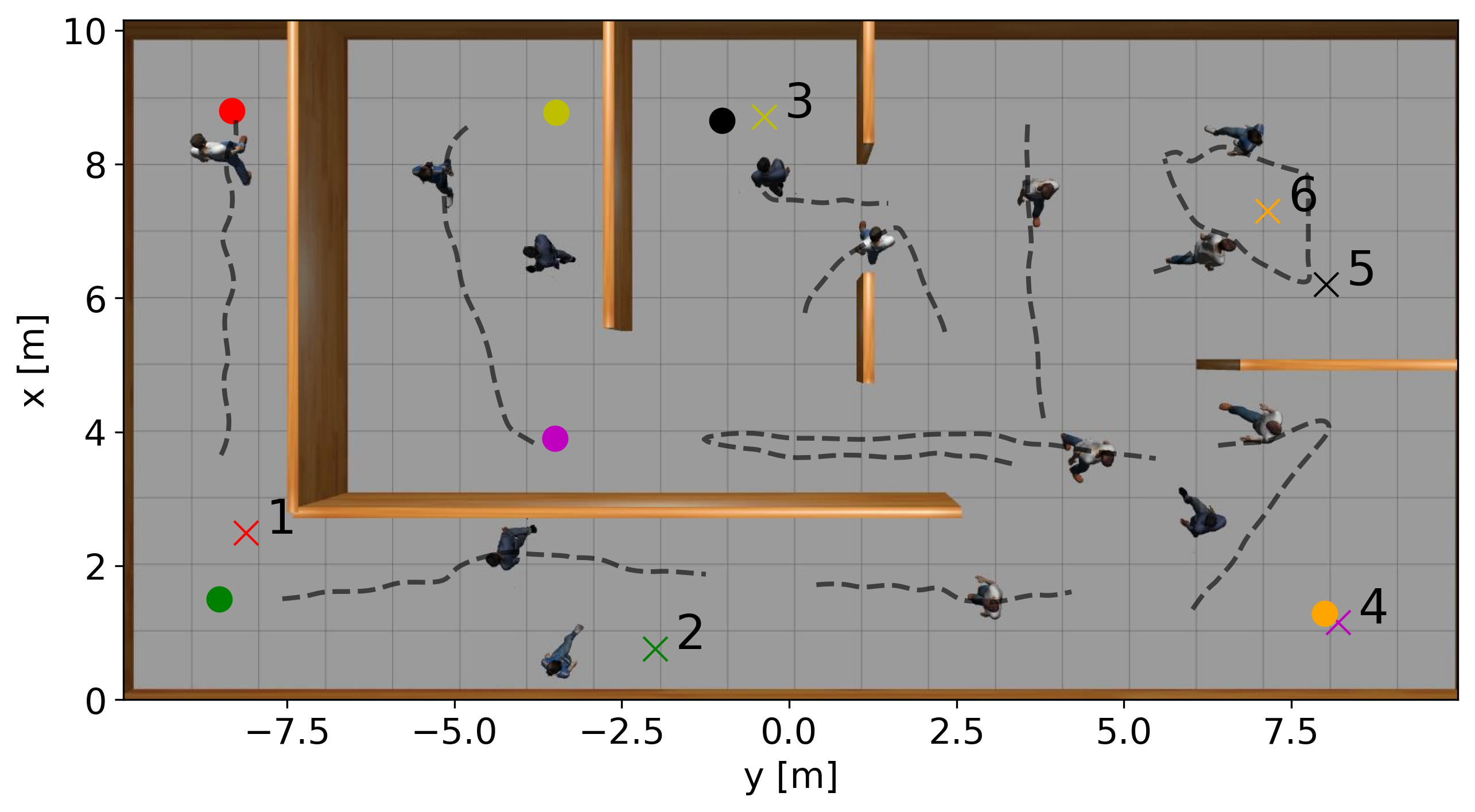}
        \label{fig:gazebo1}
    }    
    {
        \includegraphics[width=0.565\linewidth]{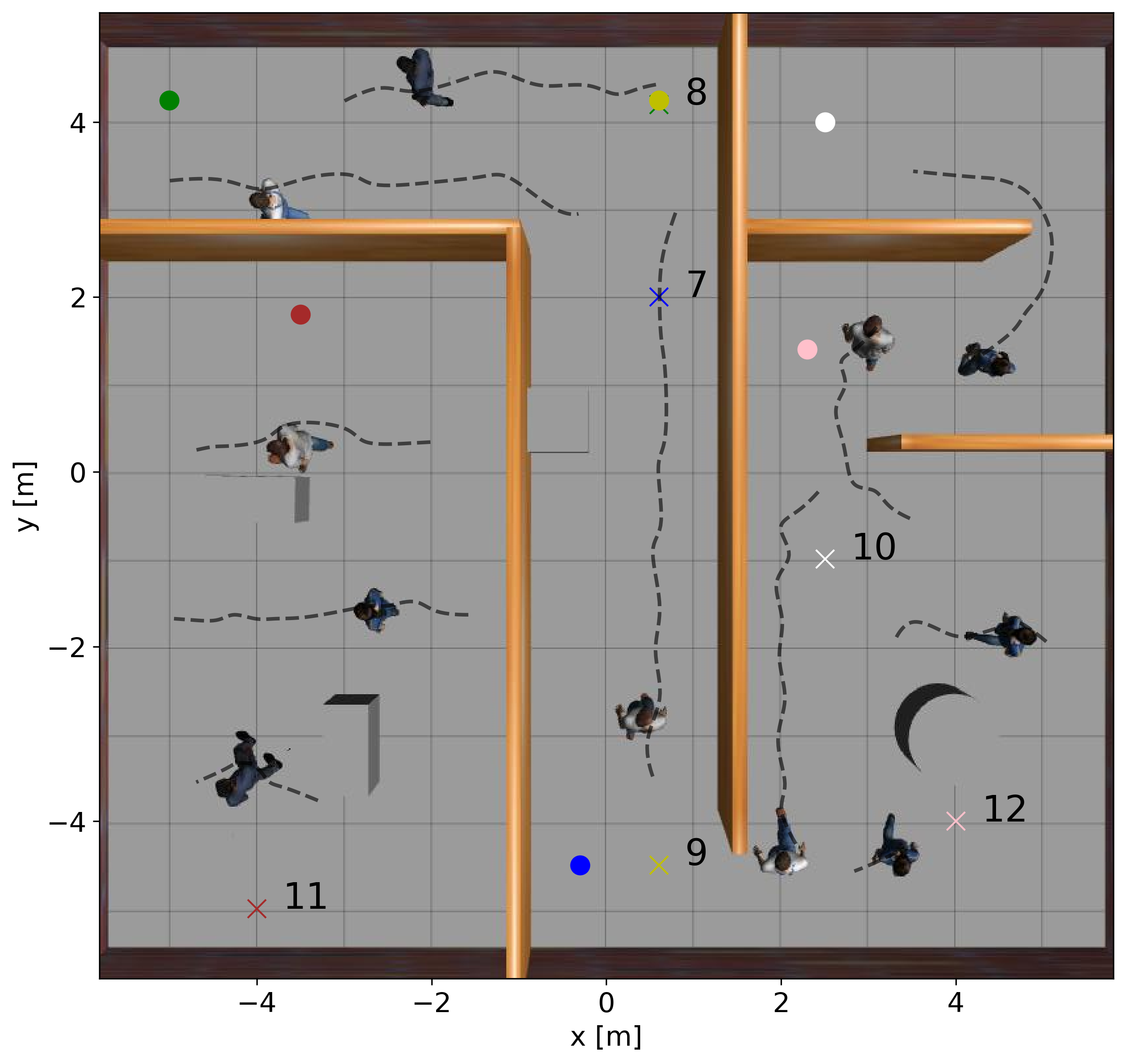}
        \label{fig:gazebo2}
    }
    \caption{Gazebo simulation environments where the agent has been trained (top) and tested (top and bottom). People trajectories are indicated with dotted lines, robot starting poses with a circle, goals with a cross and the associated episode's number.}
    \label{fig:gazebo_env}
    %Put here to reduce too much white space after your table 
\end{figure}

\begin{comment}
    
\begin{table}[ht]
\centering
\caption{Goals and robot initial poses coordinates in the testing environments shown in Figure \ref{fig:gazebo_env}.}
\label{tab:goals_and_poses_test}
\resizebox{0.8\linewidth}{!}{%
\begin{tabular}{@{}cccccc@{}}
\toprule
\multirow{2}{*}{\textbf{Env}} & \multicolumn{2}{c}{\textbf{Goal}} & \multicolumn{3}{c}{\textbf{Initial Pose}} \\ \cmidrule(l){2-6} 
                              & x [m]    & y [m]      & x [m] & y [m] & $\psi$ [rad] \\ \midrule 
1  & 2.49 & 8.110  & 8.80  & 8.320  & -3.000 \\
2  & 0.76 & 2.000  & 1.50  & 8.510  & -1.571 \\
3  & 8.70 & 0.369  & 8.77  & 3.479  & -3.139 \\
4  & 1.15 & -8.200 & 3.90  & 3.500  & -1.576 \\
5  & 6.20 & -8.020 & 8.65  & 1.000  & -2.610 \\
6  & 7.30 & -7.150 & 1.29  & -8.000 & 0.100  \\
7  & 0.60 & 2.000  & -0.30 & -4.500 & 1.570  \\
8  & 0.60 & 4.200  & -5.00 & 4.250  & 0.000  \\
9  & 0.60 & -4.500 & 0.60  & 4.250  & -1.570 \\
10 & 2.50 & -1.000 & 2.50  & 4.000  & 0.000  \\
\bottomrule
\end{tabular}%
}
\end{table}
\end{comment}

\subsubsection{State definition}
The information included in the input state of the policy network has been selected to be a synthetic but complete description of the main environmental and task-specific aspects. The state $s_t$ has been, therefore designed as the ensemble of:
\begin{itemize}
    \item Goal information: $[goal_{angle}, goal_{distance}]$ with respect to the robot expressed in polar coordinate.
    \item The set of cost weights predicted at the previous time step, $[w_{d}, w_{h}, w_{v}, w_{o}, w_{s}]_{t-1}$ to provide information about the actual state of the SFW costs used for trajectory scoring.
    \item People position and velocity information is embedded in the state for the closest $K=4$ people to the robot. Position is computed in polar coordinates $[person_{angle}, person_{distance}]$, while velocity with module and orientation, both expressed in the robot frame. People are perceived at a maximum distance of $5 m$, and if people are detected to be less than $K=4$, padding at the maximum distance is used to fill the empty input features and guarantee a constant input dimension.
    \item A set of $36$ LiDAR 2D points saturated at $3 m$ to provide the agent with the necessary awareness of local environmental geometry and spaces and the presence of obstacles.
\end{itemize}

\subsubsection{Output actions}
The policy network predicts an action $a_t = [w_d, w_h, w_v, w_o, w_s]$ at each time step, directly representing the new set of cost weights for the Social Force Window local planner. The weights are chosen in the interval of values $[0.1, 5.0]$, and they are set with a frequency of $2 Hz$, which has been considered a proper choice for a robot moving at a maximum translational velocity of $v_{max}=0.6 [m/s]$ in dynamic social scenarios.
%\end{description}

\begin{table}[ht]
\centering
\caption{Controller parameters. On the left the kinematic and classic DWA parameters, on the right the cost function weights used by the SFW controller and modified by the SFW-SAC in the range reported. The DWA uses the same cost weights except for the social term.}
\label{tab:controller_params}
\resizebox{0.7\linewidth}{!}{%
\begin{tabular}{@{}ll|lcc@{}}
\toprule
\textbf{DWA parameter} & \textbf{Value} & \textbf{Cost weight} & \textbf{SFW} & \textbf{SFW-SAC} \\ \midrule
max linear vel  & 0.6 [m/s]   & social weight   & 2.0 & [0.5 - 3.0] \\
min linear vel  & 0.08 [m/s]  & costmap weight  & 2.0 & [0.5 - 3.0] \\
max angular vel & 1.5 [rad/s] & velocity weight & 0.8 & [0.1 - 1.0] \\
waypoint tol    & 2.0 [m]     & angle weight    & 0.6 & [0.1 - 1.0] \\
sim time        & 2.5 [s]     & distance weight & 1.0 & [0.1 - 1.5] \\ \bottomrule
\end{tabular}%
}
\end{table}
\section{Experiments and Results}
\label{sec:experiments}
\subsection{Experimental settings}
The adaptive social navigation system has been trained and tested in Gazebo simulation in diverse scenarios. The HuNavSim plugin \cite{PerezRal2023} has been adopted to instantiate people moving according to the SFM with customized trajectories and behaviors; HuNavSim has also been used to collect the metrics of interest for the evaluation of the algorithms. Differently, the PIC4rl-gym \cite{martini2023pic4rl} has been used as ROS 2 framework for DRL agent training and testing.
Figure \ref{fig:gazebo_env} shows the environments realized to carry out challenging experiments categorized in different social challenges. The first Gazebo world is used for both training and testing. A wide set of diverse episodes is defined for training the agent in various conditions involving pedestrians passing, overtaking, and crossing tasks in narrow and open spaces. The agent has also been partially tested in the same world, changing the starting pose of the robot and its goals, indicated in Figure \ref{fig:gazebo_env}, scenarios $[1 - 6]$. Diversely, testing episodes $[7 - 12]$ have been performed in a separate different world to evaluate the system in diverse scenarios, always considering crossing, passing, overtaking, and mixed miscellaneous tasks.

For general and reproducible experimentation, we set a basic pedestrian behavior that considers the robot an obstacle. A global path is computed once at the beginning of each episode with the standard grid-based search planner of the Nav2 framework. The main controller parameters of the SFW algorithm are the ones of the classic DWA. Besides the kinematics limits of the robots, the waypoint position and the trajectory simulation time are important factors for a social controller, regulating the alignment to the global path and the predicting horizon. Controller parameters and cost weight values used for the experimentation are reported in Table \ref{tab:controller_params}. The static cost weights used are the results of the fine-tuning process carried out by a human expert. We use the same implementation of the SFW planner for the DWA baseline, setting the social cost to zero value. 

\begin{table}[ht]
\centering
\caption{Results obtained testing the proposed adaptive controller SFW-SAC on different environments. For each environment we report average metric results over 10 runs, comparing the agent with DWA and SFW baselines.}

\label{tab:exp_metrics}
\small
\resizebox{0.68\linewidth}{!}{%
\begin{tabular}{@{}clccccc@{}}
\toprule
\textbf{Env} & \multicolumn{1}{l}{\textbf{Method}} & \textbf{Success$\%$} & \textbf{Time {[}s{]}} & \textbf{PL {[}m{]}} & \textbf{$v_{avg}$ {[}m/s{]}} & \textbf{$SW_{step}$} \\ \midrule
\multirow{3}{*}{1}  & DWA     & 100.00 & 11.65 & 5.68  & 0.49 & 0.03 \\
                    & SFW     & 100.00 & 12.48 & 6.04  & 0.49 & 0.04 \\
                    & SFW-SAC & 100.00 & 12.21 & 5.91  & 0.48 & 0.05 \\
                    \hline
\multirow{3}{*}{2}  & DWA     & 20.00  & 12.32 & 6.21  & 0.50 & 0.17 \\
                    & SFW     & 70.00  & 20.43 & 6.42  & 0.31 & 0.11 \\
                    & SFW-SAC & 100.00 & 13.08 & 6.36  & 0.49 & 0.22 \\
                    \hline
\multirow{3}{*}{3}  & DWA     & 0.0    & -     & -     & -    & - \\
                    & SFW     & 0.0    & -     & -     & -    & - \\
                    & SFW-SAC & 70.00  & 23.26 & 11.29 & 0.48 & 0.12 \\
                    \hline
\multirow{3}{*}{4}  & DWA     & 20.00  & 21.71 & 12.21 & 0.56 & 0.25\\
                    & SFW     & 60.00  & 37.91 & 12.65 & 0.38 & 0.16\\
                    & SFW-SAC & 70.00  & 26.20 & 12.60 & 0.48 & 0.20\\
                    \hline
\multirow{3}{*}{5}  & DWA     & 50.00  & 19.53 & 9.47  & 0.49 & 0.28\\
                    & SFW     & 60.00  & 29.34 & 9.54  & 0.34 & 0.34\\
                    & SFW-SAC & 70.00  & 33.28 & 9.47  & 0.30 & 0.24\\
                    \hline
\multirow{3}{*}{6}  & DWA     & 50.00  & 19.57 & 8.87  & 0.46 & 0.26\\
                    & SFW     & 40.00  & 44.24 & 10.75 & 0.25 & 0.16\\
                    & SFW-SAC & 90.00  & 23.60 & 8.75  & 0.37 & 0.18\\
                    \hline
\multirow{3}{*}{7}  & DWA     & 0.0    & -     & -     & -    & -\\
                    & SFW     & 90.00  & 19.83 & 6.38  & 0.32 & 0.08\\
                    & SFW-SAC & 100.00 & 16.59 & 6.25  & 0.39 & 0.11\\
                    \hline
\multirow{3}{*}{8}  & DWA     & 90.00  & 10.35 & 5.21  & 0.50 & 0.10\\
                    & SFW     & 100.00 & 15.64 & 5.95  & 0.35 & 0.13\\
                    & SFW-SAC & 100.00 & 12.32 & 5.42  & 0.44 & 0.16\\
                    \hline
\multirow{3}{*}{9}  & DWA     & 80.00  & 15.75 & 8.32  & 0.53 & 0.14\\
                    & SFW     & 100.00 & 19.77 & 8.84  & 0.45 & 0.12\\
                    & SFW-SAC & 100.00 & 18.62 & 8.98  & 0.48 & 0.12\\
                    \hline
\multirow{3}{*}{10} & DWA     & 0.0    & -     & -     & -    & -\\
                    & SFW     & 70.00  & 31.96 & 8.91  & 0.29 & 0.13\\
                    & SFW-SAC & 90.00  & 32.84 & 9.99  & 0.29 & 0.14\\ 
                    \hline
\multirow{3}{*}{11} & DWA     & 50.00  & 15.45 & 6.98  & 0.45 & 0.29\\
                    & SFW     & 80.00  & 48.58 & 8.37  & 0.19 & 0.16\\
                    & SFW-SAC & 80.00  & 30.60 & 7.72  & 0.27 & 0.17\\
                    \hline
\multirow{3}{*}{12} & DWA     & 90.00  & 13.98 & 6.09  & 0.43 & 0.21\\
                    & SFW     & 40.00  & 53.04 & 6.48  & 0.14 & 0.16\\
                    & SFW-SAC & 90.00  & 32.01 & 6.29  & 0.24 & 0.20\\
                    \hline
\multirow{3}{*}{Avg}& DWA     & 58.57  & 15.84 & 8.00  & 0.50 & 0.17\\                                          & SFW     & 76.67  & 25.73 & 8.39  & 0.35 & 0.14\\
                    & SFW-SAC & 89.00  & 21.20 & 8.50  & 0.42 & 0.15\\
\bottomrule                    
\end{tabular}%
}
\end{table}

\subsection{Results}
In this section, we describe the metrics chosen to evaluate the proposed social navigation system, and we discuss the obtained results. Considering the difficulty of strictly judging the performance of a social navigation algorithm without adopting human rating, the adaptive social planner SFW-SAC has been analyzed from different perspectives. First, we compared it to the baselines DWA and Social Force Window Planner (SFW) with static costs using relevant quantitative metrics.
\begin{description}[leftmargin=0pt]
\item[Quantitative evaluation] Standard navigation metrics such as clearance time $[s]$, path length (PL) $[m]$ and average linear velocity $v_{avg} [m/s]$ are employed to evaluate the effectiveness of the planner from a classic perspective. On the other hand, the social work (SW) metric is included in the quantitative results to show the social impact of the navigation, measuring the social forces generated by the robot and on the robot by pedestrians during its motion. The social work has been taken into account as the average value $SW_{step}$ over the number of trajectory steps to consider the duration of the episode, and avoiding metrics biases caused by a fast execution of the navigation task.

A thorough inspection of the performance is presented, reporting both resulting metrics in Table \ref{tab:exp_metrics}. Results show that the DRL method enables the planner to overcome the baseline performance in multiple environments. The basic DWA fails to complete the navigation task in a high percentage of scenarios, colliding with obstacles or pedestrians.
On the other hand, the SFW baseline demonstrates an improved ability to handle social navigation tasks. Even though the cost weights combination found by a human offers safe behavior in most situations, it still presents some limitations. For example, in cluttered scenarios, SFW can be hindered by high social costs, which can cause the algorithm to get stuck. Diversely, the SFW-SAC proposes a more general performance, finding a better trade-off of costs in different situations. This advantage is proved by the higher success rate obtained in almost all the environments, sometimes being the unique solution able to complete it (Env 3). 
A more detailed analysis of results tells us that SFW-SAC often finds a compromise performance between the more aggressive DWA and the SFW with high static social cost values. This trend can be noticed by looking at the time, path length, and average velocity results. On average, the adaptive planner generally chooses higher velocities than the SFW but lower than the DWA. 
\begin{figure}[H]
\centering
\begin{adjustbox}{minipage=\linewidth,scale=0.85}
    \subfloat[DWA Env.2]{
    \includegraphics[width=0.208\linewidth]{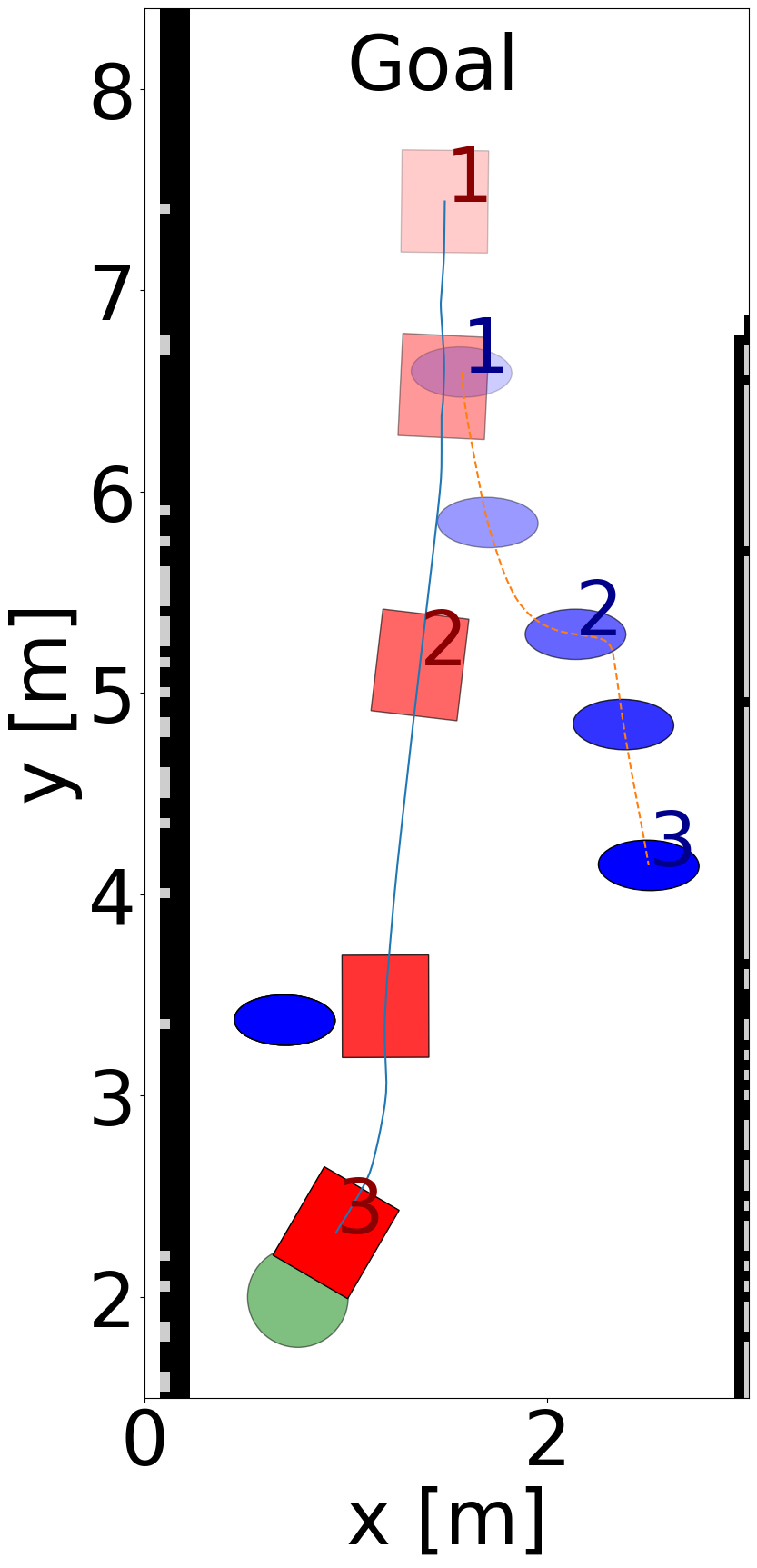}
    } 
    \subfloat[DWA Env.3]{
    \includegraphics[width=0.38\linewidth]{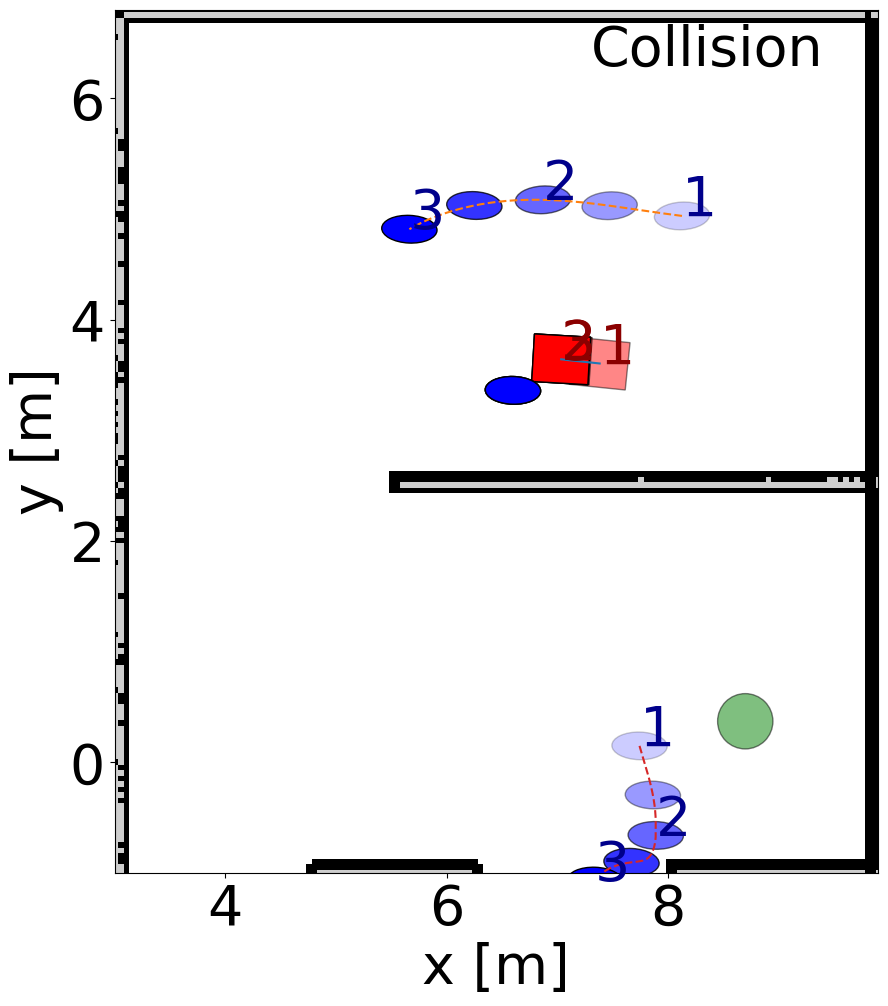}
    }
    \subfloat[DWA Env.10]{
    \includegraphics[width=0.336\linewidth]{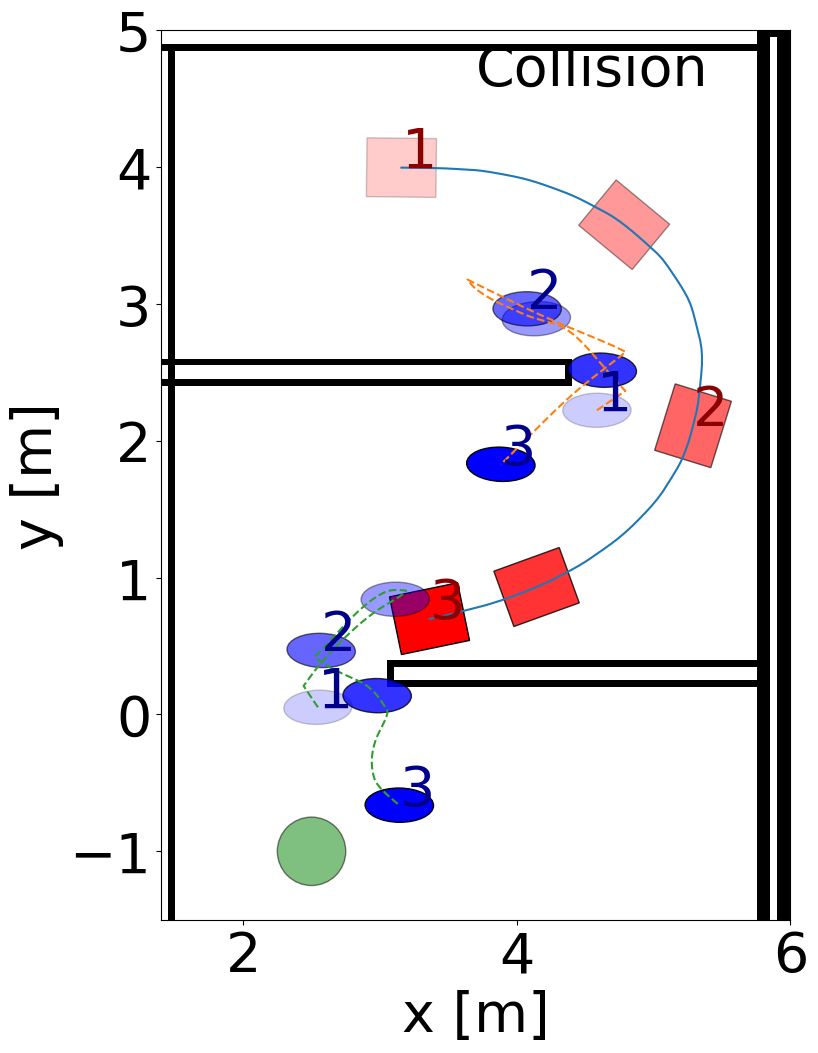}
    } 
    
    \subfloat[SFW Env.2]{
    \includegraphics[width=0.208\linewidth]{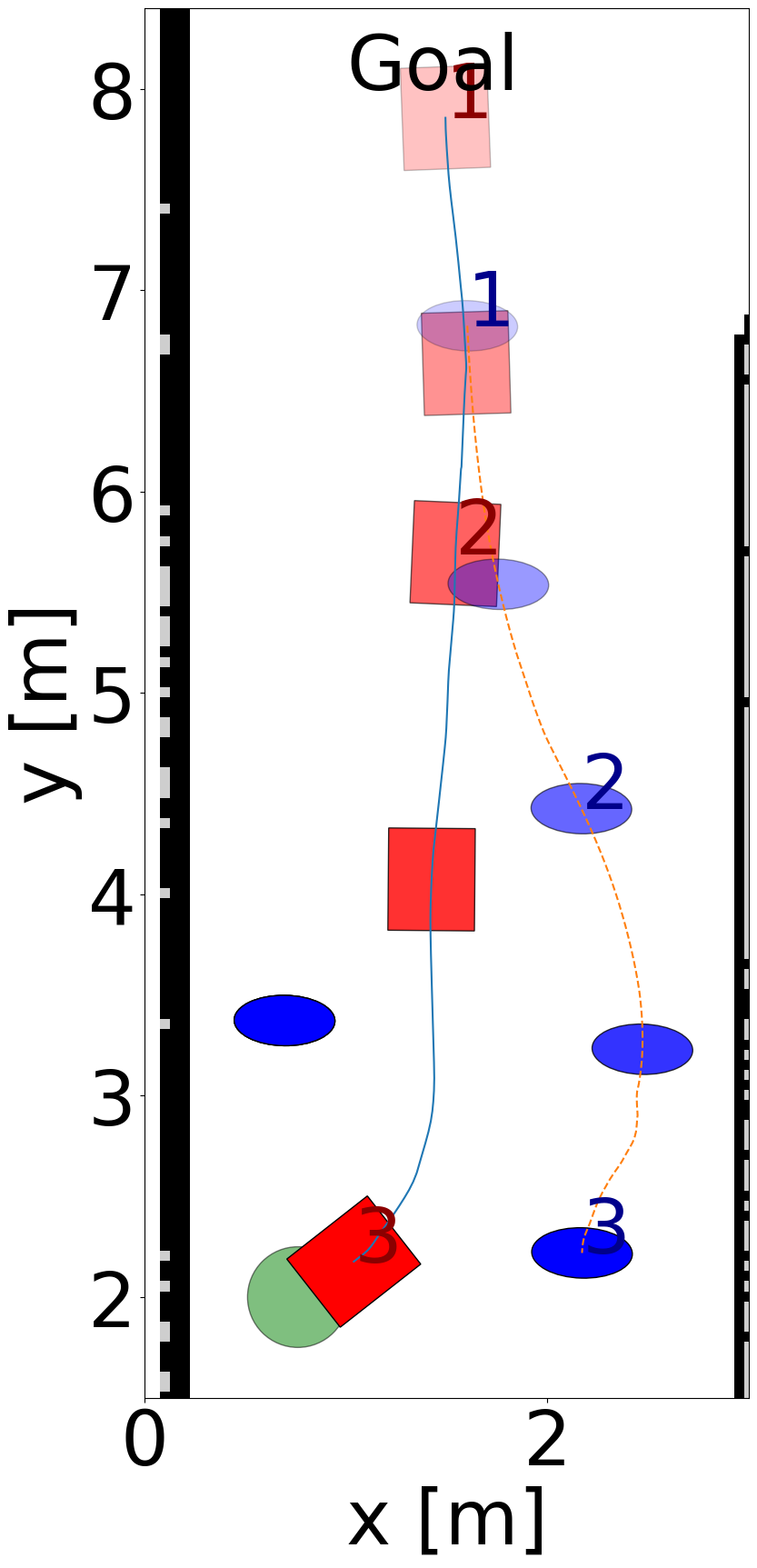}
    } 
    \subfloat[SFW Env.3]{
    \includegraphics[width=0.38\linewidth]{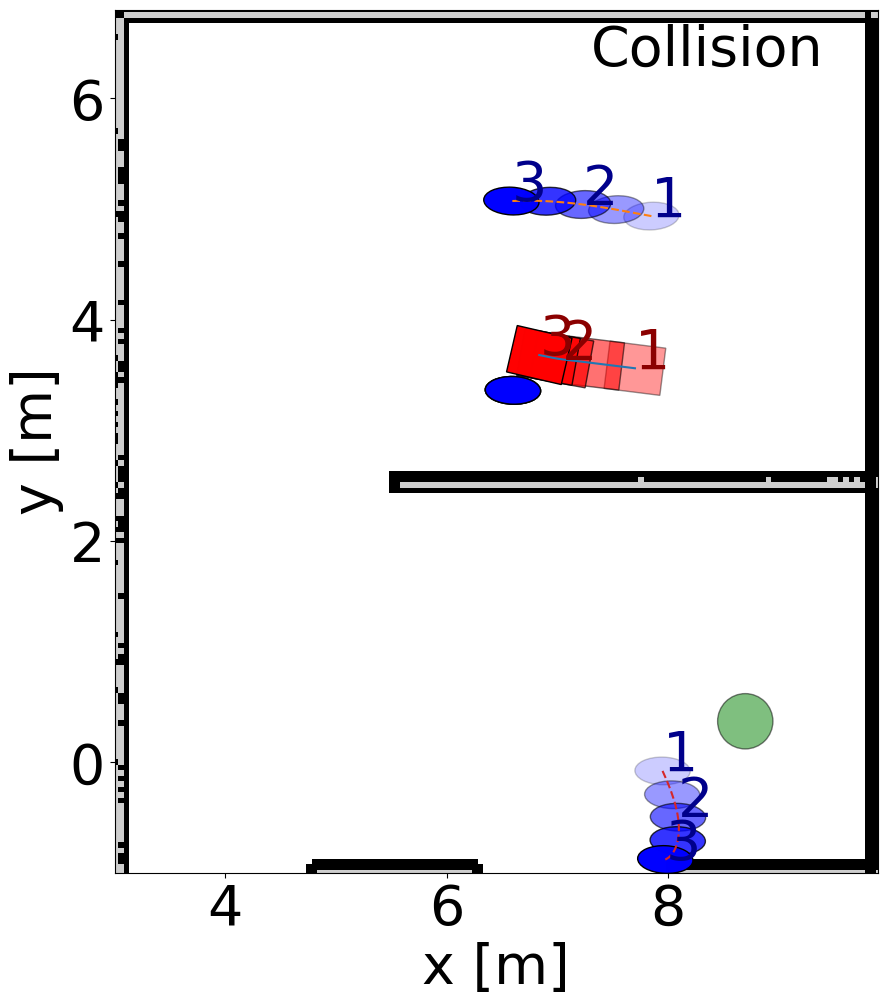}
    } 
    \subfloat[SFW Env.10]{
    \includegraphics[width=0.336\linewidth]{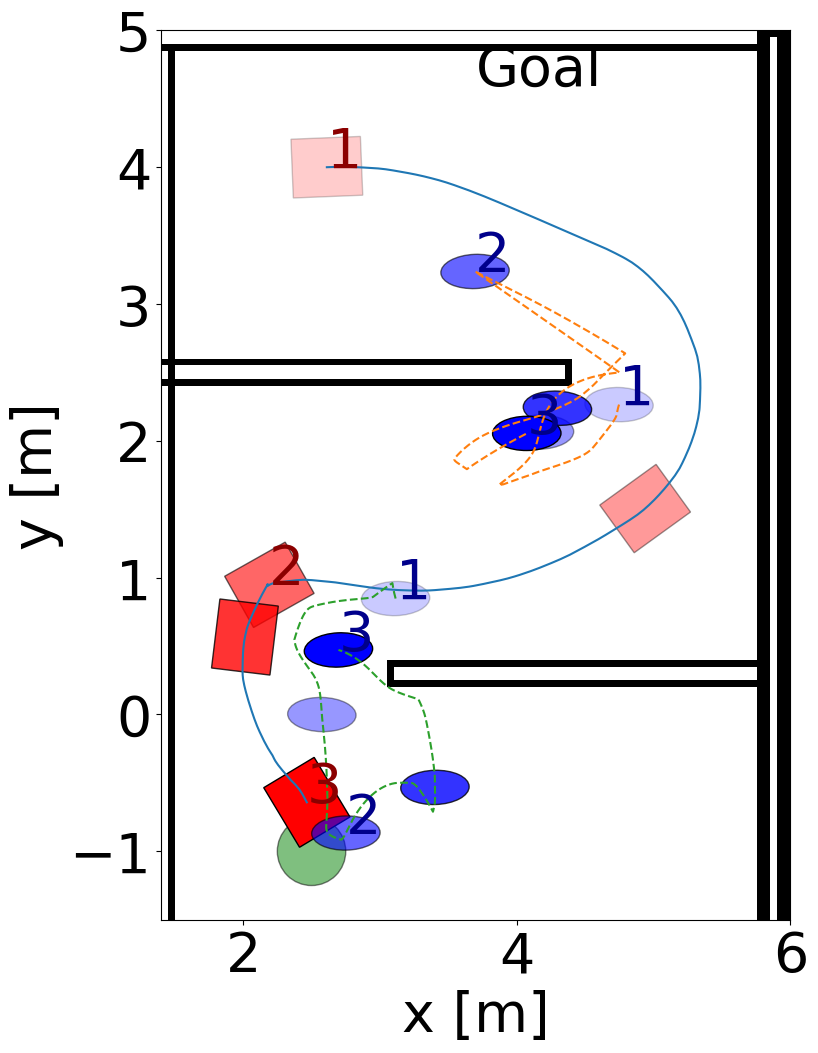}
    }

    \subfloat[SFW-SAC Env.2]{
    \includegraphics[width=0.208\linewidth]{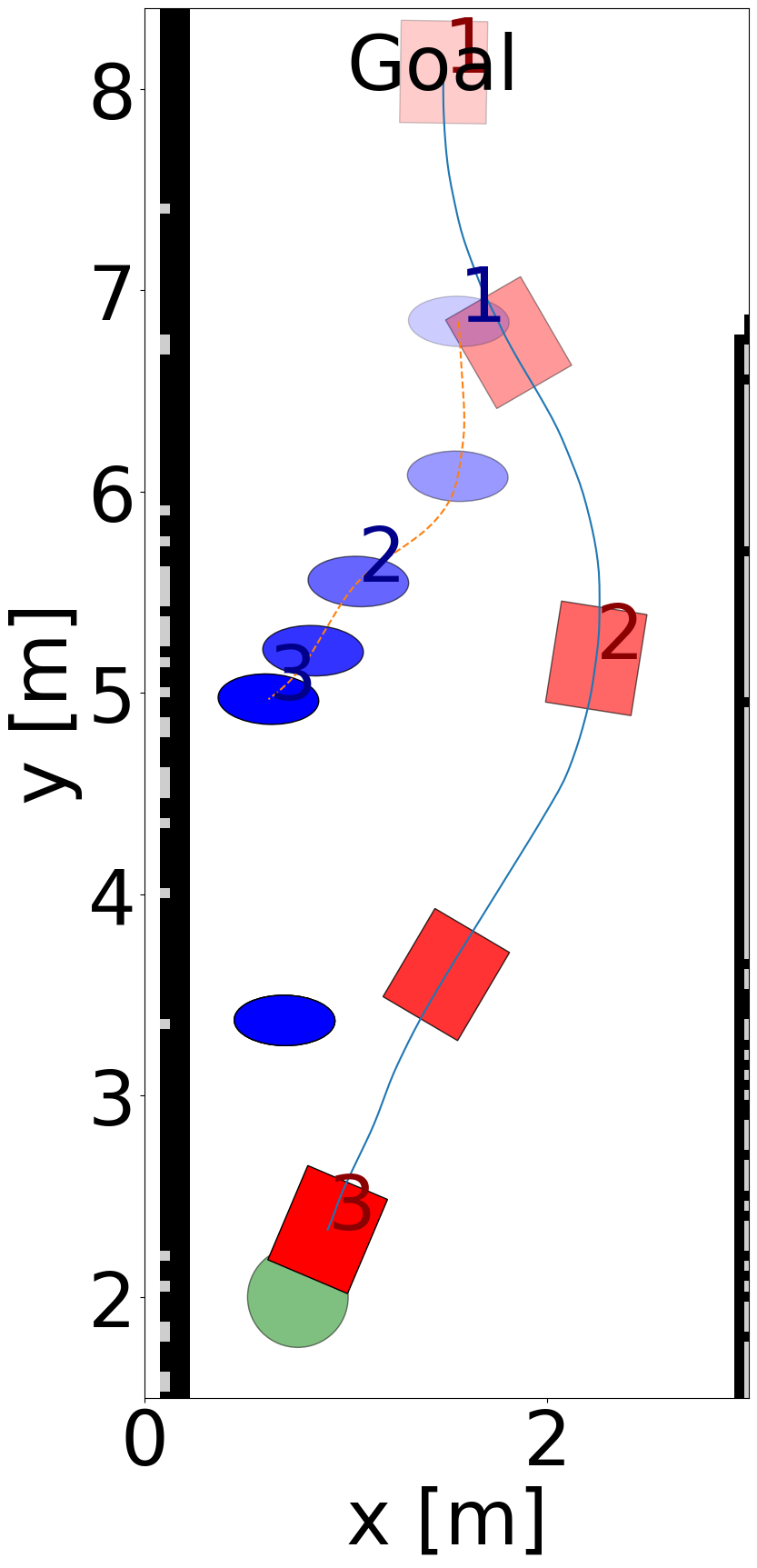}
    } 
    \subfloat[SFW-SAC Env.3]{
    \includegraphics[width=0.38\linewidth]{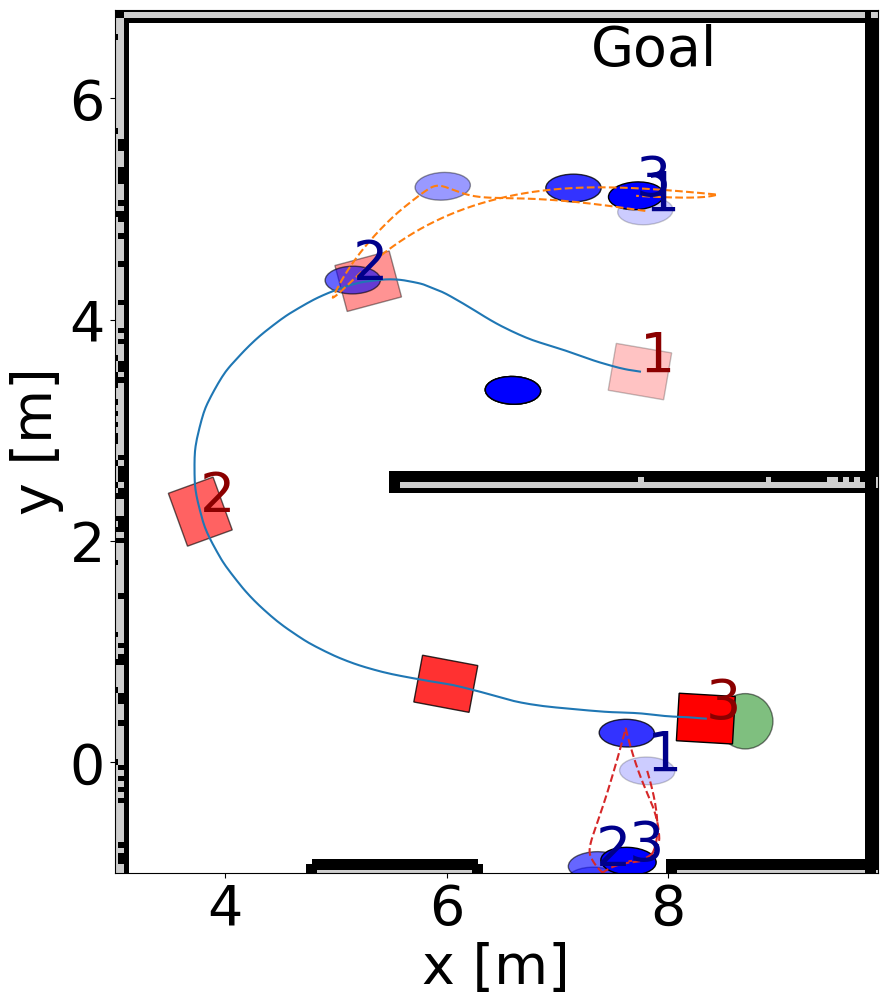}
    } 
    \subfloat[SFW-SAC Env.10]{
    \includegraphics[width=0.336\linewidth]{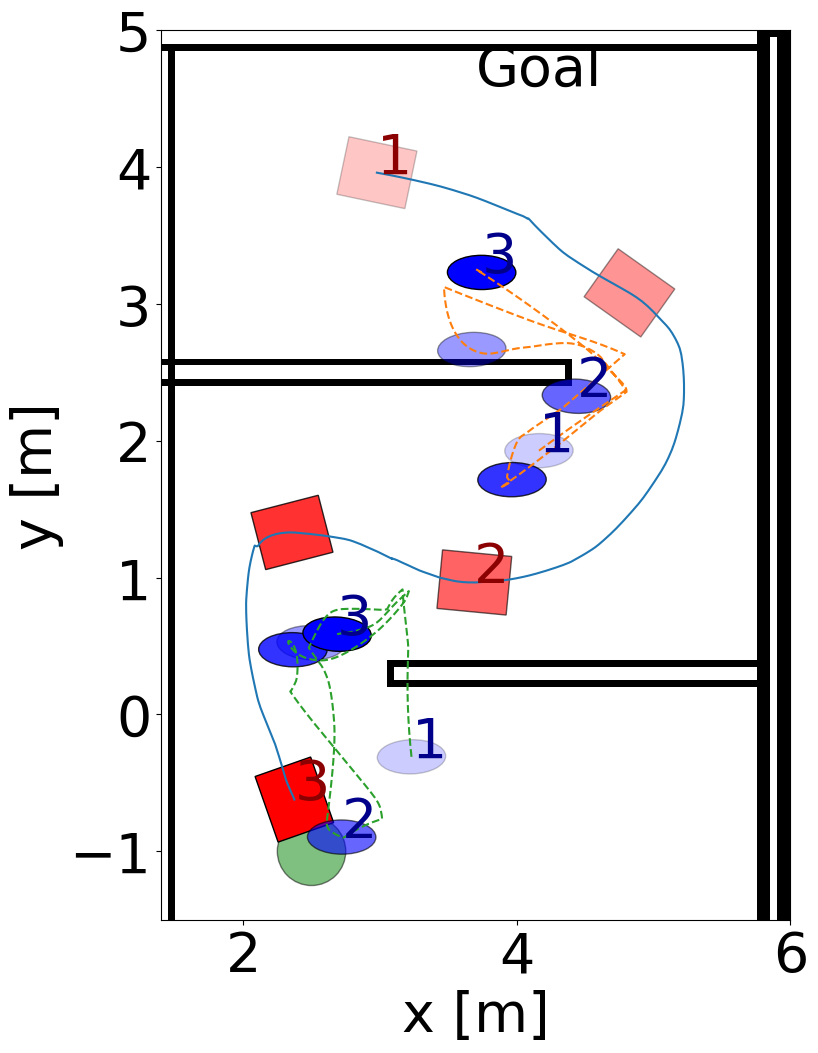}
    }
\end{adjustbox}
\caption{Trajectory plots of Env 2, 3, 10 comparing DWA, SFW and the proposed SFW-SAC adaptive planner with DRL. Goals are represented with green circles, the robot with a red rectangle and people with blue ellipses. Transparency and indexes 1,2,3 represent temporal evolution of the motion of both robot and people.} \label{fig:trajectory}
\end{figure}

Social work embeds all the navigation effects on humans, considering distances, approaching velocities, and time spent close to people. Thus, it often presents alternating results that are difficult to interpret without a visual inspection of the navigation. Indeed, DWA often reduces the duration of the episode thanks to abrupt motion and brief transitions close to people that can lead to a collision with a high risk. The $SW_{step}$, relating the social impact to the duration of the task, shows more clearly the socially compliance of SFW and SFW-SAC compared to DWA. The agent-based solution is often able to mitigate the social work improving or remaining comparable with the SFW, without compromising the success rate or strongly violating social rules.

\begin{figure}[ht]
\centering
\includegraphics[width=\linewidth]{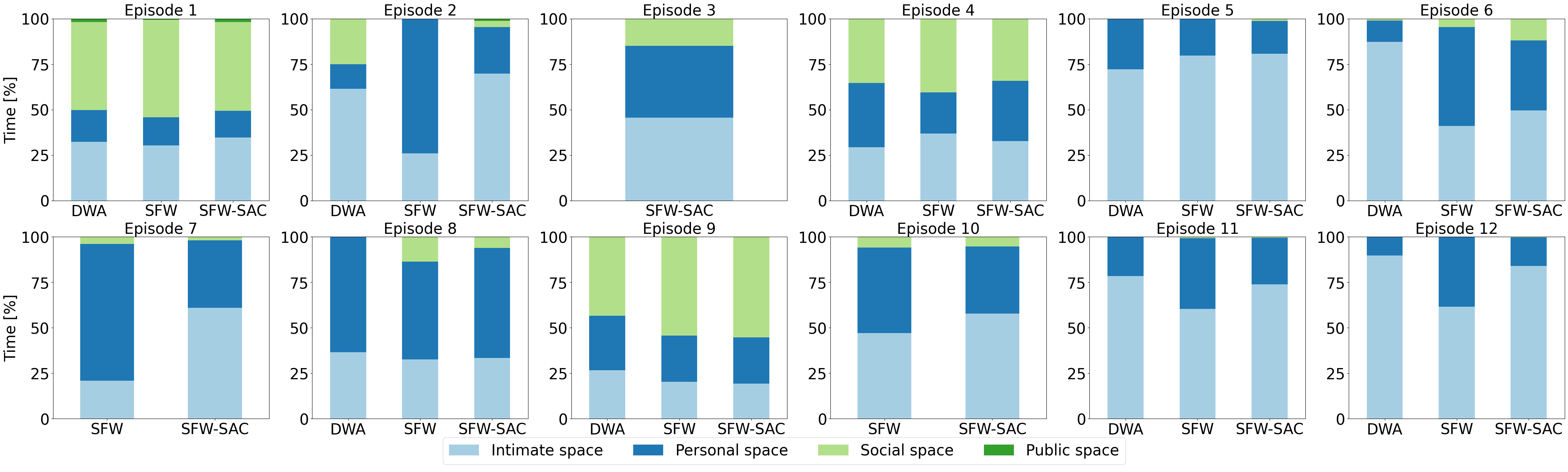}
\caption{Average temporal percentage of pedestrians space intrusion according to the proxemics standard in 10 different scenarios. Proxemics data must be coupled with success rate reported in Table \ref{tab:exp_metrics} for a clear performance frame.}
\label{fig:proxemics}
\end{figure}

% Proxemics
\item[Proxemics] According to this, the human awareness of navigation is also measured through the level of intrusion of proxemics spaces of people. Figure \ref{fig:proxemics} illustrates the percentage of time spent by the robot in the intimate, personal, social, and public space of people in each testing episode. It can be noticed that even though the SFW-SAC planner develops a more risky navigation policy, it can keep people's distances respected and comparable with the SFW baseline. It should be noted that the proxemics results reported should be paired with the success rate of the algorithms on each episode for a clear perspective. DWA often computes aggressive trajectories that do not take humans into account, although the temporal intrusion of social spaces is sometimes limited to short time intervals.
% trajectory plots of relevant scenarios
\item[Trajectory visual comparison] Resulting trajectories of some relevant scenarios where the proposed adaptive SFW-SAC show significant improvements and interesting differences, i.e., Environments $[2,3,10]$, are plotted in Figure \ref{fig:trajectory} for a better understanding of the performance of the algorithms.
In Env. 2, only the SFW-SAC can properly perform the overtaking, passing to the left of the person, while the other algorithms cross the person's path. In Env. 3, DWA and SFW are not able to handle the presence of a static person on the path and collide; the agent learns how to deviate the motion from the path and avoid the person. In Env. 10, a narrow curved passage with people passing is successfully handled by the SFW and the SFW-SAC, with a smoother trajectory.
\end{description}

\section{Conclusions}
In this work, an adaptive social planner is proposed, combining a social DWA approach with a Deep Reinforcement Learning agent. The agent boosts the performance and versatility of social navigation, learning how to adapt the controller's cost terms weights to environmental context-specific conditions. The results obtained show an improvement in both success rate and navigation metrics, balancing the trade-off between standard navigation effectiveness and social rules. Some relevant trajectories are also visualized for a clearer evaluation of the algorithms.
Future works will see the enrichment of the experimentation in different scenarios and testing the system on the real robot, including a perception system to estimate the necessary state information of the closest people to the robot.
Furthermore, the proposed method can be extended to other social controllers to be included in a common benchmark and, finally, to learning adaptive behaviors from a global planning perspective.

\subsection*{Acknowledgements} This work was partially supported by the projects NHoA (PLEC2021-007868) and NORDIC (TED2021-132476B-I00), funded by MCIN/AEI/10.13039/501100011033 and the European Union NextGenerationEU/PRTR, and partially by PoliTO Interdepartmental Centre for Service Robotics (PIC4SeR)\footnote{\url{www.pic4ser.polito.it}}.

\bibliographystyle{unsrtnat}
\bibliography{biblio}  %%% Uncomment this line and comment out the ``thebibliography'' section below to use the external .bib file (using bibtex) .





% \end{thebibliography}

\end{document}